\def\year{2021}\relax
\documentclass[letterpaper]{article} % DO NOT CHANGE THIS
\usepackage{aaai21}  % DO NOT CHANGE THIS
\usepackage{times}  % DO NOT CHANGE THIS
\usepackage{helvet} % DO NOT CHANGE THIS
\usepackage{courier}  % DO NOT CHANGE THIS
\usepackage[hyphens]{url}  % DO NOT CHANGE THIS
\usepackage{graphicx} % DO NOT CHANGE THIS
\usepackage{natbib}  % DO NOT CHANGE THIS AND DO NOT ADD ANY OPTIONS TO IT
\usepackage{caption} % DO NOT CHANGE THIS AND DO NOT ADD ANY OPTIONS TO IT
\usepackage{amsmath}
\usepackage{algorithm}
\usepackage{algorithmic}
\usepackage{multirow}
\usepackage{booktabs}
\usepackage[switch]{lineno} % add line nums

\title{Learning Modality-Specific Representations with Self-Supervised \\
Multi-Task Learning for Multimodal Sentiment Analysis}

\author{
    Wenmeng Yu,
    Hua Xu, \thanks{Corresponding Author} 
    Ziqi Yuan, 
    Jiele Wu
    \\
}
\affiliations{
    State Key Laboratory of Intelligent Technology and Systems, \\Department of Computer Science and Technology, Tsinghua University, Beijing, China\\
    Email: ywm18@mails.tsinghua.edu.cn, xuhua@mail.tsinghua.edu.cn

}

\begin{document}
% \linenumbers
\maketitle

\begin{abstract}
    Representation Learning is a significant and challenging task in 
    multimodal learning. Effective modality representations 
    should contain two parts of characteristics: the consistency and the 
    difference. Due to the unified multimodal annotation, 
    existing methods are restricted in capturing differentiated information. 
    However, additional uni-modal annotations are high time- and labor-cost. 
    In this paper, we design a label generation module based on the 
    self-supervised learning strategy to acquire independent unimodal 
    supervisions. Then, joint training the multi-modal and uni-modal tasks to learn 
    the consistency and difference, respectively. 
    Moreover, during the training stage, we design a weight-adjustment strategy 
    to balance the learning progress among different subtasks. 
    That is to  guide the subtasks to focus on samples with a larger difference 
    between modality supervisions. 
    Last, we conduct extensive experiments on three public multimodal 
    baseline datasets. The experimental results validate the reliability and 
    stability of auto-generated unimodal supervisions. 
    On MOSI and MOSEI datasets, our method surpasses 
    the current state-of-the-art methods. 
    On the SIMS dataset, our method achieves comparable performance than 
    human-annotated unimodal labels. 
    The full codes are available at \url{https://github.com/thuiar/Self-MM}.
\end{abstract}

%-------------------------------------------------------------------
\section{Introduction}
%-------------------------------------------------------------------
Multimodal Sentiment Analysis (MSA) attracts more and more attention 
in recent years \cite{TFN, MulT, poria2020beneath}. 
Compared with unimodal sentiment analysis, 
multimodal models are more robust  
and achieve salient improvements when dealing with social media data. 
With the booming of user-generated online content, 
MSA has been introduced into many applications such as risk management, 
video understanding, and video transcription.

%-------------------------------------------------------------------
% \begin{figure}[t]
%     \centering
%     \includegraphics[width=0.4\textwidth]{assets/Introduction.pdf}
%     \caption{An example of unimodal labels and multimodal 
%     labels, from \citet{TFN}. The green dotted lines represent the 
%     process of backpropagation.}
%     \label{fig: Introduction}
% \end{figure}
%-------------------------------------------------------------------

Though previous works have made impressive improvements on benchmark datasets, 
MSA is still full of challenges. 
\citet{Survey} identified five core challenges for multimodal learning: 
alignment, translation, representation, fusion, and co-learning. 
Among them, representation learning stands in a fundamental position. 
In most recent work, \citet{MISA} stated that 
unimodal representations should contain both consistent and complementary information. 
According to the difference of guidance in representation learning, we classify 
existing methods into the two categories: forward guidance and backward guidance. 
In forward-guidance methods, researches are devoted to design interactive modules for 
capturing the cross-modal information \cite{MFN, ICCN, MulT, MAG}. 
However, due to the unified multimodal annotation, 
it is difficult for them to capture modality-specific information.
% Shown in Figure \ref{fig: Introduction}, the unified multimodal labels 
% are not always suitable for the unimodal learning. 
In backward-guidance methods, researches proposed additional loss function as prior 
constraint, which leads modality representations to contain both consistent and 
complementary information\cite{SIMS,MISA}.

\citet{SIMS} introduced independent unimodal human annotations. 
By joint learning unimodal and multimodal tasks, the proposed multi-task multimodal 
framework learned modality-specific and modality-invariant representations simultaneously. 
\citet{MISA} designed two distinct encoders projecting each modality into 
modality-invariant and modality-specific space. 
Two regularization components are claimed to aid modality-invariant 
and modality-specific representation learning. 
However, in the former unimodal annotations need additional labor costs, 
and in the latter, spatial differences are difficult to 
represent the modality-specific difference.
Moreover, they require manually balanced weights between constraint components 
in the global loss function, which highly relies on the human experience.

In this paper, we focus on the backward-guidance method. 
Motivated by the independent unimodal annotations and advanced 
modality-specific representation learning, 
we propose a novel self-supervised multi-task learning strategy. 
Different from \citet{SIMS}, our method does not need human-annotated unimodal labels 
but uses auto-generated unimodal labels. 
It is based on two intuitions. 
First, label difference is positively correlated with 
the distance difference between modality representations and class centers. 
Second, unimodal labels are highly related to multimodal labels. 
Hence, we design a unimodal label generation module based on multimodal labels and 
modality representations. The details are shown in Section \ref{sec: ULGM}. 

Considering that auto-generated unimodal labels are not 
stable enough at the beginning epochs, we design a momentum-based update method, 
which applies a larger weight for the unimodal labels generated later. 
Furthermore, we introduce a self-adjustment strategy to adjust each subtask's 
weight when integrating the final multi-task loss function. 
We believe that it is difficult for subtasks with small label differences,  
between auto-generated unimodal labels and human-annotated multimodal labels,   
to learn the modality-specific representations.
Therefore, the weight of subtasks is positively correlated with the labels difference.

The novel contributions of our work can be summarized as follows:

\begin{itemize}
\item We propose the relative distance value based on the distance between modality 
    representations and class centers, positively correlated with model outputs.
\item We design a unimodal label generation module based on the self-supervised strategy. 
    Furthermore, a novel weight self-adjusting strategy is introduced 
    to balance different task loss constraints. 
\item Extensive experiments on three benchmark datasets validate the stability and reliability 
    of auto-generated unimodal labels. Moreover, our method outperforms current 
    state-of-the-art results.
\end{itemize}

%-------------------------------------------------------------------
\section{Related Work}
%-------------------------------------------------------------------
In this section, we mainly discuss related works in the domain of 
multimodal sentiment analysis and multi-task learning. 
We also emphasize the innovation of our work.
%-------------------------------------------------------------------
\subsection{Multimodal Sentiment Analysis}
%-------------------------------------------------------------------
Multimodal sentiment analysis has become a significant research topic that integrates verbal and 
nonverbal information like visual and acoustic. Previous researchers mainly focus on 
representation learning and multimodal fusion. For representation learning methods, 
\citet{RAVEN} constructed a recurrent attended variation embedding network to 
generate multimodal shifting. 
\citet{MISA} presented modality-invariant and modality-specific representations 
for representation learning in multimodal. 
For multimodal fusion, according to the fusion stage, 
previous works can be classified into two categories: early fusion and late fusion.
Early fusion methods usually use delicate attention mechanisms for cross-modal fusion. 
\citet{MFN} designed a memory fusion network for cross-view interactions. 
\citet{MulT} proposed cross-modal transformers, which learn the cross-modal attention  
to reinforce a target modality. 
Late fusion methods learn intra-modal representation first and perform inter-modal fusion last.
\citet{TFN} used a tensor fusion network that obtains tensor representation 
by computing the outer product between unimodal representations. 
\citet{LMF} proposed a low-rank multimodal fusion method to decrease the 
computational complexity of tensor-based methods.

Our work aims at representation learning based on late fusion structure. 
Different from previous studies, we joint learn unimodal and multimodal tasks 
with the self-supervised strategy. Our method learns similarity information from 
multimodal task and learns differentiated information from unimodal tasks.

\subsection{Transformer and BERT}

Transformer is a sequence-to-sequence architecture without recurrent structure \cite{attention}. 
It is used for modeling sequential data and has superior performance on results, speed, and depth 
than recurrent structure. BERT (Bidirectional Encoder Representations from Transformers) \cite{BERT}
is a successful application on the transformer. The pre-trained BERT model has 
achieved significant improvements in multiple NLP tasks. 
In multimodal learning, pre-trained BERT also achieved remarkable results. 
Currently, there are two ways to use pre-trained BERT. 
The first is to use the pre-trained BERT as a language feature extraction module \cite{MISA}.
The second is to integrate acoustic and visual information on the 
middle layers \cite{MulT, MAG}. In this paper, we use the first way and finetune
the pre-trained BERT for our tasks.

\subsection{Multi-task Learning}

Multi-task learning aims to improve the generalization performance of multiple related 
tasks by utilizing the knowledge contained in different tasks \cite{Multi-task}.
Compared with single-task learning, there are two main challenges for 
multi-task learning in the training stage. 
The first is how to share network parameters, including hard-sharing and soft-sharing 
methods. The second is how to balance the learning process of different tasks.
Recently, multi-task learning is wildly applied in MSA\cite{liu2015multi, zhang2016deep, akhtar2019multi, CH-SIMS}. 

In this work, we introduce unimodal subtasks to aid the modality-specific representation learning.
We adopt a hard-sharing strategy and design a weight-adjustment method to solve 
the problem of how to balance.

%-------------------------------------------------------------------
\begin{figure*}[t]
    \centering
    \includegraphics[width=0.8\textwidth]{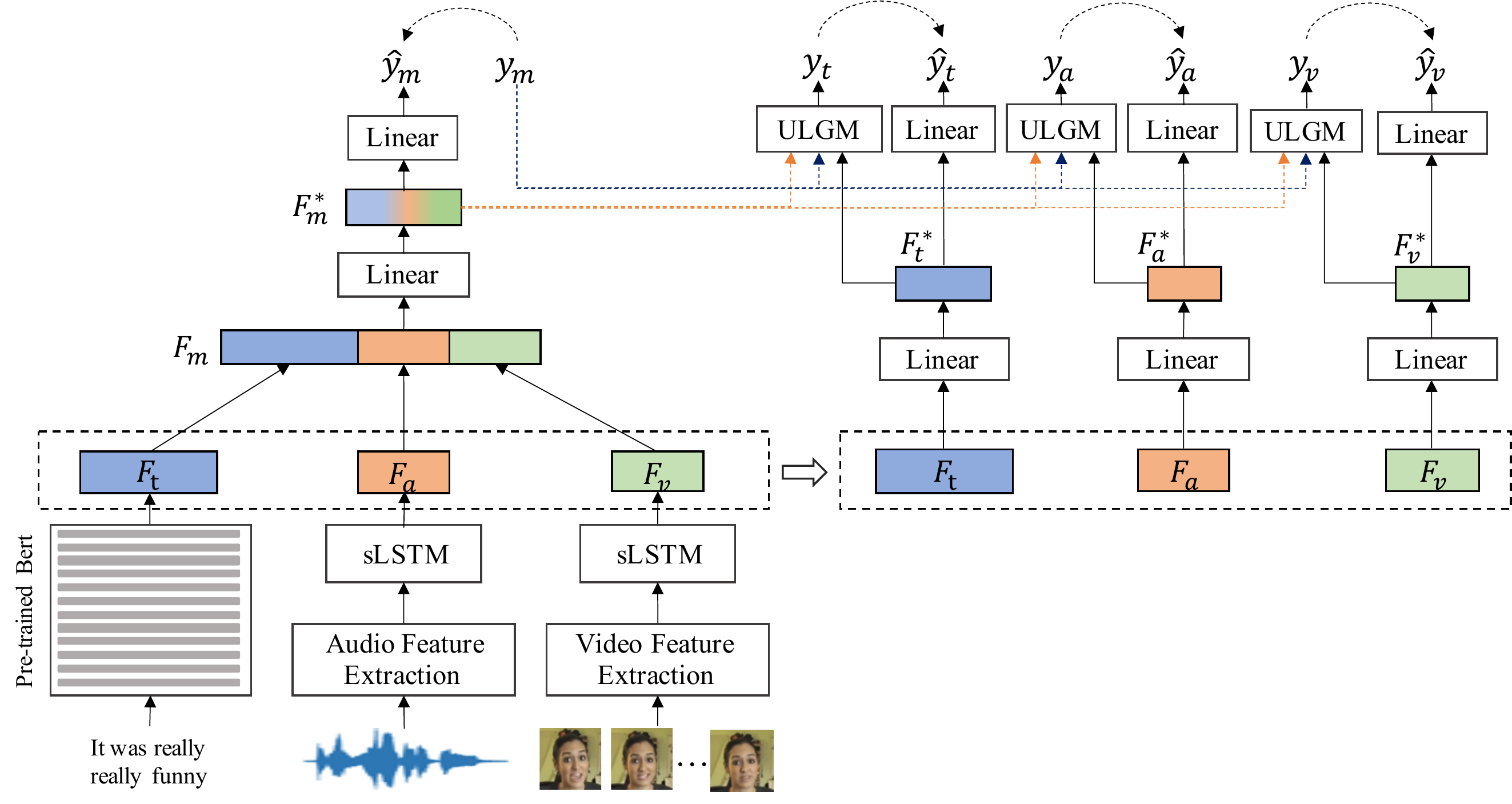}
    \caption{The overall architecture of Self-MM. 
    The $\hat{y}_m$, $\hat{y}_t$, $\hat{y}_a$, and $\hat{y}_v$ are 
    the predictive outputs of the multimodal task and the three unimodal tasks, respectively. 
    The $y_m$ is the multimodal annotation by human. 
    The $y_t$, $y_a$, and $y_v$ are the unimodal supervision 
    generated by the self-supervised strategy. Finally, $\hat{y}_m$ is used 
    as the sentiment output.}
    \label{fig: MainModel}
\end{figure*}
%-------------------------------------------------------------------
%-------------------------------------------------------------------
\section{Methodology}
%-------------------------------------------------------------------
In this section, we explain the Self-Supervised Multi-task Multimodal 
sentiment analysis network (Self-MM) in detail. 
The goal of the Self-MM is to acquire information-rich unimodal 
representations by joint learning one multimodal task and 
three unimodal subtasks. Different from the multimodal task, 
the labels of unimodal subtasks are auto-generated in the self-supervised method.
For the convenience of the following sections, we refer the human-annotated 
multimodal labels as \textbf{m-labels} and the auto-generated unimodal labels 
as \textbf{u-labels}.

\subsection{Task Setup}
Multimodal Sentiment Analysis (MSA) is to judge the sentiments 
using multimodal signals, including text ($I_t$), audio ($I_a$), and vision ($I_v$). 
Generally, MSA can be regarded as either a regression task or a classification task. 
In this work, we regard it as the regression task. 
Therefore, Self-MM takes $I_t, I_a, \text{and} I_v$ as inputs and 
outputs one sentimental intensity result $\hat{y}_m \in R$.
In the training stage, to aid representation learning, 
Self-MM has extra three unimodal outputs $\hat{y}_s \in R$, where $s \in \{ t, a, v\}$. 
Though more than one output, we only use $\hat{y}_m$ as the final predictive result.

%-------------------------------------------------------------------
\subsection{Overall Architecture}
%-------------------------------------------------------------------
Shown in Figure \ref{fig: MainModel}, the Self-MM consists of 
one multimodal task and three independent unimodal subtasks. Between the 
multimodal task and different unimodal tasks, we adopt hard-sharing 
strategy to share the bottom representation learning network.

\noindent \textbf{Multimodal Task.} 
For the multimodal task, we adopt a classical multimodal sentiment 
analysis architecture. It contains three main parts: 
the feature representation module, the feature fusion module, and the output module. 
In the text modality, since the great success of the pre-trained language model, 
we use the pre-trained 12-layers BERT to extract sentence representations. 
Empirically, the first-word vector in the last layer is selected as the whole sentence representation $F_t$.  

\begin{equation*}
    F_t = BERT(I_t; \theta_t^{bert}) \in R^{d_t}
\end{equation*}

In audio and vision modalities, following \citet{TFN, CH-SIMS}, 
we use pre-trained ToolKits to extract the initial vector features, 
$I_a \in R^{l_a \times d_a}$ and $I_v \in R^{l_v \times d_v}$, from raw data. 
Here, $l_a$ and $l_v$ are the sequence lengths of audio and vision, respectively. 
Then, we use a single directional Long Short-Term Memory (sLSTM) \cite{LSTM} 
to capture the timing characteristics. Finally, the end-state hidden vectors are 
adopted as the whole sequence representations. 

\begin{align*}
    F_a &= sLSTM(I_a; \theta_a^{lstm}) \in R^{d_a}\\
    F_v &= sLSTM(I_v; \theta_v^{lstm}) \in R^{d_v}
\end{align*}

Then, we concatenate all uni-modal representations and 
project them into a lower-dimensional space $R^{d_m}$.

\begin{equation*}
    F_m^* = ReLU({W_{l1}^m}^T [F_t; F_a; F_v] + b_{l1}^m)
\end{equation*} 
\noindent where $W_{l1}^m \in R^{(d_t+d_a+d_v) \times d_m}$ and 
$ReLU$ is the relu activation function.

Last, the fusion representation $F_m^*$ is used to predict the 
multimodal sentiment. 
\begin{equation*}
    \hat{y}_m = {W_{l2}^m}^T F_m^* + b_{l2}^m
\end{equation*}
\noindent where $W_{l2}^m \in R^{d_m \times 1}$.

\noindent \textbf{Uni-modal Task.} For the three unimodal tasks, 
they share modality representations with the multimodal task.  
In order to reduce the dimensional difference between different 
modalities, we project them into a new feature space. 
Then, get the uni-modal results with linear regression.  
\begin{equation*}
    F_s^* = ReLU({W_{l1}^s}^T F_s + b_{l1}^s) \\
\end{equation*}
\begin{equation*}
    \hat{y}_s = {W_{l2}^s}^T F_s^* + b_{l2}^s
\end{equation*}
\noindent where $s \in \{t, a, v\}$.

To guide the unimodal tasks' training process, 
we design a Unimodal Label Generation Module (ULGM) to get u-labels. 
Details of the ULGM are discussed in Section \ref{sec: ULGM}.  

\begin{equation*}
    y_s = ULGM(y_m, F_m^*, F_s^*)
\end{equation*}
\noindent where $s \in \{t, a, v\}$.

Finally, we joint learn the multimodal task and three unimodal tasks under 
m-labels and u-labels supervision.
It is worth noting that these unimodal tasks are only exist in 
the training stage. Therefore, we use $\hat{y}_m$ as the final output.

% TO-DO
%-------------------------------------------------------------------
\subsection{ULGM}
\label{sec: ULGM}
%-------------------------------------------------------------------
\begin{figure}[t]
    \centering
    \includegraphics[width=0.4\textwidth]{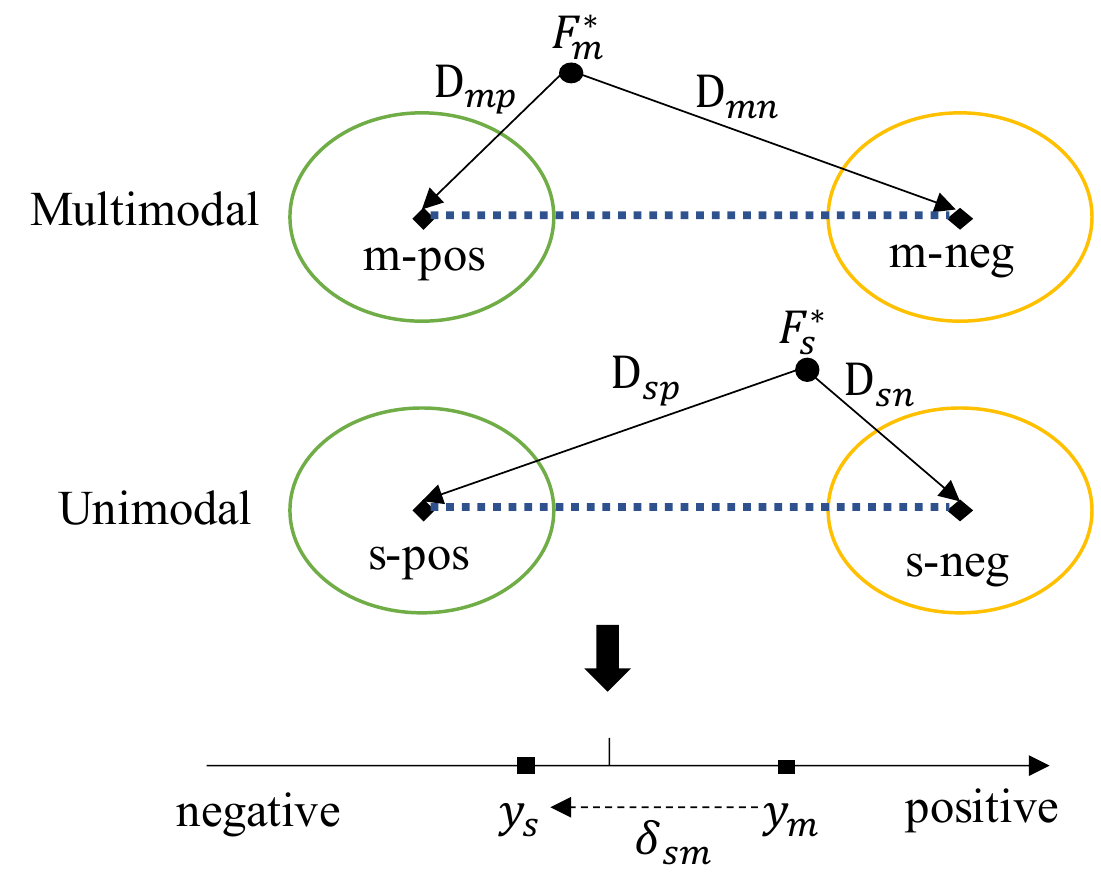}
    \caption{Unimodal label generation example. Multimodal representation 
    $F_m^*$ is closer to the positive center (m-pos) while 
    unimodal representation is closer to the negative center (s-neg). 
    Therefore, unimodal supervision $y_s$ is 
    added a negative offset $\delta_{sm}$ to the multimodal label $y_m$}.
    \label{fig: DistExp}
\end{figure}

% ------------------------------------------------------------------------------
\begin{algorithm}[t]
    \renewcommand{\algorithmicrequire}{\textbf{Input:}}
	\renewcommand{\algorithmicensure}{\textbf{Output:}}
    \caption{Unimodal Supervisions Update Policy}
    \label{alg: updateLabel}
    \begin{algorithmic}[1]
        \REQUIRE unimodal inputs $I_t, I_a, I_v$, m-labels $y_m$
        \ENSURE u-labels $y_t^{(i)}, y_a^{(i)}, y_v^{(i)}$ where $i$ means the number of training epochs
        \STATE Initialize model parameters $M(\theta;x)$
        \STATE Initialize u-labels $y_t^{(1)} = y_m, y_a^{(1)} = y_m, y_v^{(1)} = y_m$
        \STATE Initialize global representations $F_t^g = 0, F_a^g = 0, F_v^g = 0, F_m^g = 0$
        \FOR {$n \in [1, end]$}
            \FOR {mini-batch in dataLoader}
                \STATE Compute mini-batch modality representations $F_t^*, F_a^*, F_v^*, F_m^*$
                \STATE Compute loss $L$ using Equation~(\ref{eqn: loss})
                \STATE Compute parameters gradient $\frac{\vartheta L}{\vartheta \theta}$
                \STATE Update model parameters: $\theta = \theta - \eta \frac{\vartheta L}{\vartheta \theta}$
                \IF {$n \neq 1$}
                    \STATE Compute relative distance values $\alpha_m, \alpha_t, \alpha_a,$ and $\alpha_v$ using 
                    Equation~(\ref{eqn: classCenter}$\sim$\ref{eqn: rdv})
                    \STATE Compute $y_t, y_a, y_v$ using Equation~(\ref{eqn: computeLabel})
                    \STATE Update $y_t^{(n)}, y_a^{(n)}, y_t^{(n)}$ using Equation~(\ref{eqn: updateLabel})
                \ENDIF
                \STATE Update global representations $F_s^g$ using $F_s^*$, where $s \in \{m,t,a,v\}$
            \ENDFOR
        \ENDFOR
    \end{algorithmic}
\end{algorithm}
% ------------------------------------------------------------------------------
%-------------------------------------------------------------------
\begin{table*}[t]
    \centering
    \begin{tabular}{cccccccccc}
    % \hline
    \toprule[1pt]
    \multirow{2}{*}{Model} & \multicolumn{4}{c}{MOSI}                  & \multicolumn{4}{c}{MOSEI}                 & \multirow{2}{*}{\begin{tabular}[c]{@{}c@{}}Data\\ Setting\end{tabular}} \\ 
    \cmidrule(r){2-5} \cmidrule(r){6-9}
                    & MAE            & Corr           & Acc-2                & F1-Score             & MAE            & Corr           & Acc-2                 & F1-Score             &                                               \\ 
    \midrule[1pt]
    TFN (B)$^1$     & 0.901          & 0.698          & -/80.8               & -/80.7               & 0.593          & 0.700          & -/82.5                & -/82.1               & Unaligned                                     \\
    LMF (B)$^1$     & 0.917          & 0.695          & -/82.5               & -/82.4               & 0.623          & 0.677          & -/82.0                & -/82.1               & Unaligned                                     \\ 
    MFN$^1$         & 0.965          & 0.632          & 77.4/-               & 77.3/-               & -              & -              & 76.0/-                & 76.0/-               & Aligned                                       \\
    RAVEN$^1$       & 0.915          & 0.691          & 78.0/-               & 76.6/-               & 0.614          & 0.662          & 79.1/-                & 79.5/-               & Aligned                                       \\
    MFM (B)$^1$     & 0.877          & 0.706          & -/81.7               & -/81.6               & 0.568          & 0.717          & -/84.4                & -/84.3               & Aligned                                       \\
    MulT (B)$^1$    & 0.861          & 0.711          & 81.5/84.1            & 80.6/83.9            & 0.58           & 0.703          & -/82.5                & -/82.3               & Aligned                                       \\
    MISA (B)$^1$    & 0.783          & 0.761          & 81.8/83.4            & 81.7/83.6            & 0.555          & 0.756          & 83.6/85.5             & 83.8/85.3            & Aligned                                       \\
    MAG-BERT (B)$^2$& 0.712          & 0.796          & 84.2/86.1            & 84.1/86.0            & -              & -              & 84.7/-                & 84.5/-               & Aligned                                       \\
    \hline
    MISA (B)*       & 0.804          & 0.764          & 80.79/82.1           & 80.77/82.03          & 0.568          & 0.724          & 82.59/84.23           & 82.67/83.97          & Aligned                                       \\
    MAG-BERT (B)*   & 0.731          & 0.789          & 82.54/84.3           & 82.59/84.3           & 0.539          & 0.753          & \textbf{83.79}/\textbf{85.23}  & \textbf{83.74}/85.08 & Aligned                                       \\ 
    Self-MM (B)*    & \textbf{0.713} & \textbf{0.798} & \textbf{84.00/85.98} & \textbf{84.42/85.95} & \textbf{0.530} & \textbf{0.765} & 82.81/85.17  & 82.53/\textbf{85.30} & Unaligned                                     \\ 
    \bottomrule[1pt]
    \end{tabular}
    \caption{Results on MOSI and MOSEI. (B) means the language features are 
    based on BERT; $^1$ is from \citet{MISA} and $^2$ is 
    from \citet{MAG}. Models with $^*$ are reproduced under the same 
    conditions. In Acc-2 and F1-Score, the left of the ``/'' is calculated as 
    ``negative/non-negative'' and the right is calculated as ``negative/positive''.}
    \label{tab: res-mosi-mosei}
\end{table*}
%-------------------------------------------------------------------
\noindent The ULGM aims to generate uni-modal supervision values based on 
multimodal annotations and modality representations. In order to avoid 
unnecessary interference with the update of network parameters, the ULGM 
is designed as a non-parameter module. 
Generally, unimodal supervision values are highly correlated 
with multimodal labels. Therefore, the ULGM calculates the offset 
according to the relative distance from modality representations to 
class centers, shown as Figure \ref{fig: DistExp}. 

\noindent \textbf{Relative Distance Value.} 
Since different modality representations exist in different feature spaces, 
using the absolute distance value is not accurate enough.  
Therefore, we propose the relative distance value, which is not related to  
the space difference.  First, when in training process, 
we maintain the positive center ($C_i^p$) and the negative center ($C_i^n$) of different modality representations: 
\begin{eqnarray}
    \label{eqn: classCenter}
    C_i^p &= \frac{\sum_{j=1}^{N} I(y_i(j) > 0) \cdot F_{ij}^g}{\sum_{j=1}^{N} I(y_i(j) > 0)} \\
    C_i^n &= \frac{\sum_{j=1}^{N} I(y_i(j) < 0) \cdot F_{ij}^g}{\sum_{j=1}^{N} I(y_i(j) < 0)} 
\end{eqnarray}
\noindent where $i \in \{m, t, a, v\}$, $N$ is the number of training samples, and 
$I(\cdot)$ is a indicator function. $F_{ij}^g$ is the  global representation 
of the $j_{th}$ sample in modality $i$.

For modality representations, we use L2 normalization 
as the distance between $F_{i}^*$ and class centers. 
\begin{eqnarray}
    D_{i}^p &= \frac{||F_{i}^* - C_i^p||_2^2}{\sqrt{d_i}}\\
    D_{i}^n &= \frac{||F_{i}^* - C_i^n||_2^2}{\sqrt{d_i}}
\end{eqnarray}
\noindent where $i \in \{ m, t, a, v\}$. $d_i$ is the representation dimension, 
a scale factor.

Then, we define the relative distance value, which evaluates the relative distance 
from the modality representation to the positive center and the negative center. 
\begin{eqnarray}
    \alpha_{i} = \frac{D_{i}^n - D_{i}^p}{D_{i}^p + \epsilon}
    \label{eqn: rdv}
\end{eqnarray}
\noindent where $i \in \{ m, t, a, v\}$. $\epsilon$ is a small number in case of zero exceptions. 

\noindent \textbf{Shifting Value.} 
It is intuitive that $\alpha_{i}$ is positively related to 
the final results. To get the link between supervisions and predicted values, 
we consider the following two relationships. 
\begin{equation}
    \frac{y_s}{y_m} \propto \frac{\hat{y}_s}{\hat{y}_m} \propto \frac{\alpha_s}{\alpha_m} 
    \Rightarrow y_s = \frac{\alpha_s * y_m}{\alpha_m}\\
    \label{eqn: div}
\end{equation}
\begin{equation}
    y_s - y_m \propto \hat{y}_s - \hat{y}_m \propto \alpha_s - \alpha_m 
    \Rightarrow y_s = y_m + \alpha_s - \alpha_m
    \label{eqn: minus}
\end{equation}
\noindent where $s \in \{t, a, v\}$.

Specifically, the Equation \ref{eqn: minus} is introduced to avoid the ``zero value problem''. 
In Equation \ref{eqn: div}, when $y_m$ equals to zero, the generated unimodal supervision 
values $y_s$ are always zero. Then, joint considering the above relationships, 
we can get unimodal supervisions by equal-weight summation.
\begin{align}
    \begin{split}
    y_s &= \frac{y_m * \alpha_s}{2 \alpha_m} + \frac{y_m + \alpha_s - \alpha_m}{2} \\
    &= y_m + \frac{\alpha_s - \alpha_m}{2} * \frac{y_m + \alpha_m}{\alpha_m}\\
    &= y_m + \delta_{sm}
    \end{split}
    \label{eqn: computeLabel}
\end{align}
\noindent where $s \in \{ t, a, v \}$. The $\delta_{sm} = \frac{\alpha_t - \alpha_m}{2} * \frac{y_m + \alpha_m}{\alpha_m}$ 
represents the offset value of unimodal supervisions to multimodal annotations. 

\noindent\textbf{Momentum-based Update Policy.}
Due to the dynamic changes of modality representations, the generated 
u-labels calculated by Equation (\ref{eqn: computeLabel}) are unstable enough. 
In order to mitigate the adverse effects, we design a momentum-based update policy,  
which combines the new generated value with history values.
\begin{eqnarray}
    y_s^{(i)} = 
    \begin{cases}
    y_m & i = 1 \\
    \frac{i-1}{i+1} y_s^{(i-1)} + \frac{2}{i+1} y_s^{i} & i > 1
    \end{cases}
    \label{eqn: updateLabel}
\end{eqnarray}
\noindent where $s \in \{ t, a, v \}$. $y_s^i$ is the new generated u-labels at 
the $i_{th}$ epoch. $y_s^{(i)}$ is the final u-labels after the $i_{th}$ epoch. 

Formally, assume the total epochs is $n$, we can get that the weight of $y_s^i$ 
is $\frac{2i}{(n)(n+1)}$. It means that the weight of u-labels generated later is 
greater than the previous one. It is in accordance with our experience. 
Because generated unimodal labels are the cumulative sum of all previous epochs, 
they will stabilize after enough iterations (about 20 in our experiments). 
Then, the training process of unimodal tasks will gradually become stable.
The unimodal labels update policy is shown in Algorithm \ref{alg: updateLabel}. 

%-------------------------------------------------------------------
\subsection{Optimization Objectives}
%-------------------------------------------------------------------
Finally, we use the L1Loss as the basic optimization objective. For uni-modal 
tasks, we use the difference between u-labels and m-labels as the weight of 
loss function. It indicates that 
the network should pay more attention on the samples with larger difference. 
\begin{eqnarray}
    L = \frac{1}{N} \sum_i^N (|\hat{y}_m^i - y_m^i| + 
        \sum_s^{\{t,a,v\}} W_s^i * |\hat{y}_s^i - y_s^{(i)}|)
    \label{eqn: loss}
\end{eqnarray}
\noindent where $N$ is the number of training samples. 
$W_s^i = tanh(|y_s^{(i)} - y_m|)$ is the weight of $i_{th}$ sample for 
auxiliary task $s$.

%-------------------------------------------------------------------
\section{Experimental Settings}
%-------------------------------------------------------------------
In this section, we introduce our experimental settings, including 
experimental datasets, baselines, and evaluations.
%-------------------------------------------------------------------
\subsection{Datasets}
%-------------------------------------------------------------------
%-------------------------------------------------------------------
\begin{table}[t]
    \centering
    \begin{tabular}{ccccc}
    \toprule[1pt]
    Dataset & \# Train & \# Valid & \# Test & \# All \\ 
    \midrule[1pt]
    MOSI    & 1284     & 229      & 686     & 2199   \\
    MOSEI   & 16326    & 1871     & 4659    & 22856  \\
    SIMS    & 1368     & 456      & 457     & 2281   \\ 
    \bottomrule[1pt]
    \end{tabular}
    \caption{Dataset statistics in MOSI, MOSEI, and SIMS.}
    \label{tab: stats}
\end{table}
%-------------------------------------------------------------------
%-------------------------------------------------------------------
\begin{figure*}[t]
    \centering
    \includegraphics[width=0.8\textwidth]{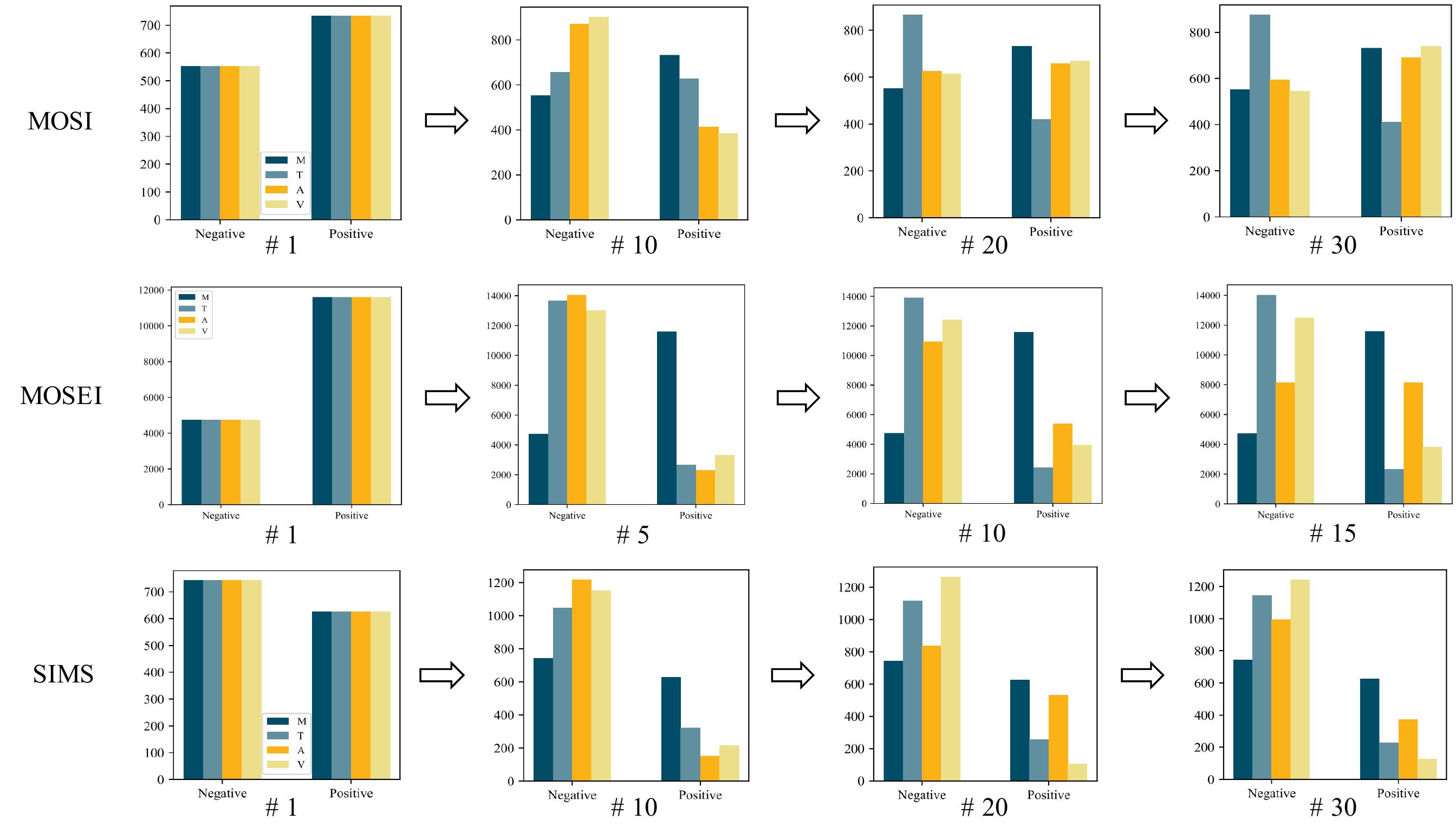}
    \caption{The distribution update process of u-labels on different 
    datasets. The number (\#) under each sub picture indicates the number of epochs.}
    \label{fig: label-distribution}
\end{figure*}
%-------------------------------------------------------------------

In this work, we use three public multimodal sentiment analysis datasets, 
MOSI \cite{MOSI}, MOSEI \cite{MOSEI}, and SIMS \cite{SIMS}. The basic statistics 
are shown in Table \ref{tab: stats}. Here, we give a brief introduction to 
the above datasets.

\noindent \textbf{MOSI.} The CMU-MOSI dataset \cite{MOSI} is one of the most 
popular benchmark datasets for MSA. It comprises 2199 short monologue 
video clips taken from 93 Youtube movie review videos. Human annotators 
label each sample with a sentiment score from -3 (strongly negative) 
to 3 (strongly positive).

\noindent \textbf{MOSEI.} The CMU-MOSEI dataset \cite{MOSEI} expands its 
data with a higher number of utterances, greater variety in samples, speakers, 
and topics over CMU-MOSI. The dataset contains 23,453 annotated video 
segments (utterances), from 5,000 videos, 1,000 distinct speakers and 
250 different topics. 

\noindent \textbf{SIMS.} The SIMS dataset\cite{SIMS} is a distinctive Chinese 
MSA benchmark with fine-grained annotations of modality. The dataset 
consists of 2,281 refined video clips collected from different movies, 
TV serials, and variety shows with spontaneous expressions, various head 
poses, occlusions, and illuminations. Human annotators label each sample with 
a sentiment score from -1 (strongly negative) to 1 (strongly positive).

%-------------------------------------------------------------------
\subsection{Baselines}
%-------------------------------------------------------------------
To fully validate the performance of the Self-MM, we make a fair comparison with 
the following baselines and state-of-the-art models in multimodal sentiment analysis. 

\noindent \textbf{TFN.} The Tensor Fusion Network (TFN) \cite{TFN} calculates 
a multi-dimensional tensor (based on outer-product) to capture uni-, bi-, 
and tri-modal interactions.

\noindent \textbf{LMF.} The Low-rank Multimodal Fusion (LMF) \cite{LMF} 
is an improvement over TFN, where low-rank multimodal tensors fusion 
technique is performed to improve efficiency.

\noindent \textbf{MFN.} The Memory Fusion Network (MFN) \cite{MFN} 
accounts for continuously modeling the view-specific and cross-view 
interactions and summarizing them through time with a Multi-view Gated Memory.

\noindent \textbf{MFM.} The Multimodal Factorization Model (MFM) \cite{MFM} 
learns generative representations to learn the modality-specific 
generative features along with discriminative representations for 
classification.

\noindent \textbf{RAVEN.} The Recurrent Attended Variation Embedding 
Network (RAVEN) \cite{RAVEN} utilizes an attention-based model 
re-adjusting word embeddings according to auxiliary non-verbal signals.

\noindent \textbf{MulT.} The Multimodal Transformer (MulT) \cite{MulT} 
extends multimodal transformer architecture with directional pairwise 
crossmodal attention which translates one modality to another using 
directional pairwise cross-attention.

\noindent \textbf{MAG-BERT.} The Multimodal Adaptation Gate for Bert 
(MAG-BERT) \cite{MAG} is an improvement over RAVEN on aligned 
data with applying multimodal adaptation gate at different layers of 
the BERT backbone.

\noindent \textbf{MISA.} The Modality-Invariant and -Specific 
Representations (MISA) \cite{MISA} incorporate a combination of 
losses including distributional similarity, orthogonal loss, reconstruction 
loss and task prediction loss to learn modality-invariant and 
modality-specific representation.

%-------------------------------------------------------------------
\begin{table}[t]
    \centering
    \begin{tabular}{ccccc}
    \toprule[1pt]
    Model                          & MAE   & Corr  & Acc-2 & F1-Score \\ 
    \midrule[1pt]
    TFN                            & 0.428 & 0.605 & 79.86 & 80.15    \\
    LMF                            & 0.431 & 0.600 & 79.37 & 78.65    \\ 
    \hline
    \multicolumn{1}{l}{Human-MM}   & 0.408 & 0.647 & 81.32 & 81.73    \\
    Self-MM                        & 0.419 & 0.616 & 80.74 & 80.78    \\ 
    \bottomrule[1pt]
    \end{tabular}
    \caption{Results on SIMS.}
    \label{tab: res-sims}
\end{table}
%-------------------------------------------------------------------
%-------------------------------------------------------------------
\subsection{Basic Settings}
%-------------------------------------------------------------------
\noindent \textbf{Experimental Details.}
We use Adam as the optimizer and use the initial learning rate of $5e-5$ for Bert and 
$1e-3$ for other parameters. For a fair comparison, in our model (Self-MM) and 
two state-of-the-art methods (MISA and MAG-BERT), we run five times 
and report the average performance.

\noindent \textbf{Evaluation Metrics.}
Following the previous works \cite{MISA,MAG}, we report our experimental results 
in two forms: classification and regression. For classification, we report 
Weighted F1 score (F1-Score) and binary classification accuracy (Acc-2). 
Specifically, for MOSI and MOSEI datasets, we calculate Acc-2 and F1-Score 
in two ways: negative / non-negative (non-exclude zero)\cite{TFN} 
and negative / positive (exclude zero)\cite{MulT}.
For regression, we report Mean Absolute Error (MAE) and Pearson correlation (Corr). 
Except for MAE, higher values denote better performance for all metrics. 

%-------------------------------------------------------------------
\section{Results and Analysis}
%-------------------------------------------------------------------
In this section, we make a detailed analysis and discussion about our 
experimental results. 
%-------------------------------------------------------------------
\subsection{Quantitative Results} 
%-------------------------------------------------------------------
Table \ref{tab: res-mosi-mosei} shows the comparative results on MOSI and 
MOSEI datasets. For a fair comparison, according to the difference of ``Data Setting'', 
we split models into two categories: Unaligned and Aligned.
Generally, models using aligned corpus can get better results \cite{MulT}. 
In our experiments, first, comparing with unaligned models (TFN and LMF), 
we achieve a significant improvement in all evaluation metrics. 
Even comparing with aligned models, our method gets competitive results. 
Moreover, we reproduce the two best baselines ``MISA'' and ``MAG-BERT'' under 
the same conditions. We find that our model surpasses them on most of the evaluations.

Since the SIMS dataset only contains unaligned data, 
we compare the Self-MM with TFN and LMF. 
Besides, we use the human-annotated unimodal labels to replace 
the auto-generated u-labels, called Human-MM. Experimental results are 
shown in Table \ref{tab: res-sims}. We can find that the Self-MM gets 
better results than TFN and LMF and achieve comparable performance with 
Human-MM. The above results show that 
our model can be applied to different data scenarios and 
achieve significant improvements.

%-------------------------------------------------------------------
\begin{table}[t]
    \centering
    \begin{tabular}{ccccc}
    \toprule[1pt]
    Tasks      & MSE            & Corr           & Acc-2                & F1-Score             \\
    \hline
    M          & 0.730          & 0.781          & 82.38/83.67          & 82.48/83.70          \\
    M, V       & 0.732          & 0.775          & 82.67/83.52          & 82.76/83.55          \\
    M, A       & 0.728          & 0.790          & 82.80/84.76          & 82.85/84.75          \\
    M, T       & 0.731          & 0.789          & 82.65/84.15          & 82.66/84.10          \\
    M, A, V    & 0.719          & 0.789          & 82.94/84.76          & 83.05/84.81          \\
    M, T, V    & 0.714          & 0.797          & \textbf{84.26}/85.91 & 84.33/\textbf{86.00} \\
    M, T, A    & \textbf{0.712} & 0.797          & 83.67/85.06          & 83.72/85.06          \\
    M, T, A, V & 0.713          & \textbf{0.798} & 84.00/\textbf{85.98} & \textbf{84.42}/85.95 \\
    \bottomrule[1pt]
    \end{tabular}
    \caption{Results for multimodal sentiment analysis with different tasks using Self-MM. 
    M, T, A, V represent the multimodal, text, audio, and vision task, respectively.}
    \label{tab: ablation}
\end{table}
%-------------------------------------------------------------------

%-------------------------------------------------------------------
\subsection{Ablation Study}
\label{sec: ablation-study}
%-------------------------------------------------------------------
To further explore the contributions of Self-MM, 
we compare the effectiveness of combining different uni-modal tasks. 
Results are shown in Table \ref{tab: ablation}. 
Overall, compared with the single-task model, 
the introduce of unimodal subtasks can significantly 
improve model performance. 
From the results, we can see that ``M, T, V'' and ``M, T, A'' achieve 
comparable or even better results than ``M, T, A, V''. 
Moreover, we can find that subtasks, ``T'' and ``A'', help more 
than the subtask ``V''. 

%-------------------------------------------------------------------
\subsection{Case Study}
\label{sec: case-study}
%-------------------------------------------------------------------
To validate the reliability and reasonability of auto-generated u-labels, 
we analyze the update process of u-labels, shown in Figure \ref{fig: label-distribution}. 
We can see that as the number of iterations increases, 
the distributions of u-labels tends to stabilize. It is in line 
with our expectations. Compared with MOSI and SIMS datasets, the 
update process on the MOSEI has faster convergence. It shows 
that the larger dataset has more stable class centers, which is more 
suitable for self-supervised methods. 

\begin{figure}[t]
    \centering
    \includegraphics[width=0.45\textwidth]{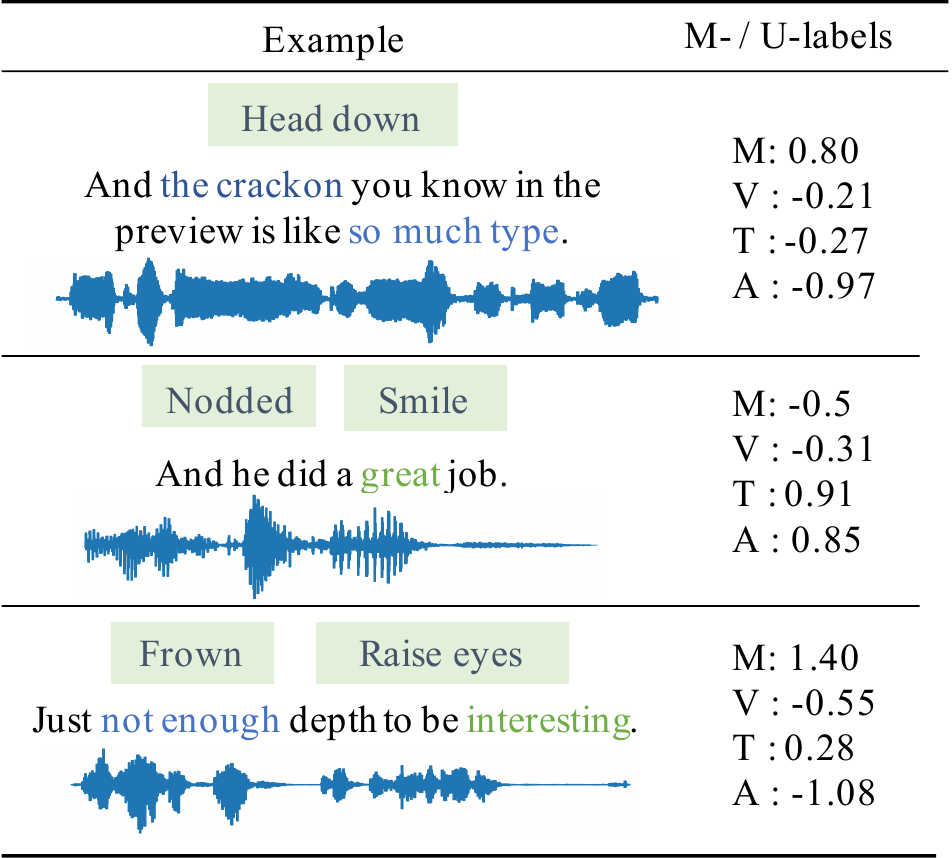}
    \caption{Case study for the Self-MM on MOSI. The ``M'' is human-annotated, 
    and ``V, T, A'' are auto-generated.}
    \label{fig: res-example}
\end{figure}

In order to further show the reasonability of the u-labels, 
we selected three multimodal examples from the MOSI dataset, 
as shown in Figure \ref{fig: res-example}. In the first and 
third cases, human-annotations m-labels are $0.80$ and $1.40$. 
However, for single modalities, they are inclined to negative sentiments. 
In line with expectation, the u-labels get negative offsets on the m-labels. 
A positive offset effect is achieved in the second case. 
Therefore, the auto-generated u-labels are significant. 
We believe that these independent u-labels can aid in learning  
modality-specific representation. 

%-------------------------------------------------------------------
\section{Conclusion and Future Work}
%-------------------------------------------------------------------
In this paper, we introduce unimodal subtasks to aid in learning  
modality-specific representations. Different from previous works, 
we design a unimodal label generation strategy based on 
the self-supervised method, which saves lots of human costs. 
Extensive experiments validate the reliability and stability of 
the auto-generated unimodal labels. We hope this work can provide a 
new perspective on multimodal representation learning. 

We also find that the generated audio and vision labels 
are not significant enough limited by the pre-processed features.
In future work, we will build an end-to-end multimodal learning network 
and explore the relationship between unimodal and multimodal learning.

%-------------------------------------------------------------------
\section{ Acknowledgments}
%-------------------------------------------------------------------
% This paper is founded by National Natural Science Foundation of China (Grant No: 61673235) 
% and National Key R\&D Program Projects of China (Grant No: 2018YFC1707605). 
% We would like to thank the anonymous reviewers for their valuable suggestions.
This paper is supported by National Key R\&D Program Projects of China (Grant No: 2018YFC1707605) 
and seed fund of Tsinghua University (Department of Computer Science and Technology) -Siemens Ltd., 
China Joint Research Center for Industrial Intelligence and Internet of Things. 
We would like to thank the anonymous reviewers for their valuable suggestions.
%-------------------------------------------------------------------
\bibliography{iyuge2}

\end{document}